\documentclass[11pt]{article}
\usepackage[top=1in, bottom=1in, left=1in, right=1in]{geometry}
\usepackage{tikz}
\usetikzlibrary{shapes,arrows,matrix,patterns,positioning}
\usepackage{color}
\usepackage{epstopdf}
\usepackage{graphicx}
\usepackage{algorithmic}
\usepackage{amssymb}
\usepackage{epstopdf}
\usepackage[norelsize,boxed,linesnumbered,vlined]{algorithm2e}
\usepackage{float}
\usepackage{amsmath,amsthm,bm,color,epsfig,enumerate,caption}
\usepackage{hyperref}
\usepackage{booktabs}
\usepackage{multirow}
\usepackage{array}
\usepackage{todonotes}
\usepackage{xspace}
\usepackage{subfig}

\usepackage{ulem}

\newcommand{\BEA}{\begin{eqnarray}}
\newcommand{\EEA}{\end{eqnarray}}
\newcommand{\comment}[1]{}
\usepackage{amssymb,amsmath,txfonts}
\newtheorem{theorem}{Theorem}

\newtheorem{property}[theorem]{Property}

\def\eg{\emph{e.g}}

\def\ie{\emph{i.e}}

\makeatletter
\newcommand*{\extendadd}{
  \mathbin{
    \mathpalette\extend@add{}
  }
}
\newcommand*{\extend@add}[2]{
  \ooalign{
    $\m@th#1\leftrightarrow$%
    \vphantom{$\m@th#1\updownarrow$}
    \cr
    \hfil$\m@th#1\updownarrow$\hfil
  }
}
\makeatother

\begin{document}

\title{Drop-Activation: Implicit Parameter Reduction and Harmonious Regularization}
\author{Senwei Liang \\ Department of Mathematics\\ Purdue University, IN 47907, USA\\ \href{mailto:liang339@purdue.edu}{liang339@purdue.edu}
        \and Yuehaw Khoo\\ Department of Statistics and the College\\ The University of Chicago, Chicago, IL 60637\\ \href{mailto:ykhoo@galton.uchicago.edu}{ykhoo@galton.uchicago.edu}
         \and Haizhao Yang \\ Department of Mathematics\\ Purdue University, IN 47907, USA\footnote{Current institute.}\\Department of Mathematics\\ National University of Singapore, Singapore\footnote{Institute when the project was started.}\\
          \href{mailto:haizhao@purdue.edu}{haizhao@purdue.edu} }

\maketitle

\begin{abstract}
Overfitting frequently occurs in deep learning. In this paper, we propose a novel regularization method called Drop-Activation to reduce overfitting and improve generalization. The key idea is to drop nonlinear activation functions by setting them to be identity functions randomly during training time. During testing, we use a deterministic network with a new activation function to encode the average effect of dropping activations randomly. Our theoretical analyses support the regularization effect of Drop-Activation as implicit parameter reduction and verify its capability to be used together with Batch Normalization~\cite{bn}. The experimental results on CIFAR-10, CIFAR-100, SVHN, EMNIST, and ImageNet show that Drop-Activation generally improves the performance of popular neural network architectures for the image classification task. Furthermore, as a regularizer Drop-Activation can be used in harmony with standard training and regularization techniques such as Batch Normalization and Auto Augment~\cite{autoaugment}. The code is available at \url{https://github.com/LeungSamWai/Drop-Activation}.
\end{abstract}


\section{Introduction}
Convolution neural network (CNN) is a powerful tool for computer vision tasks. With the help of gradually increasing depth and width, CNNs \cite{resnet, preresnet, densenet, wrn, resnext} gain a significant improvement in image classification problems by capturing multiscale features \cite{visual}. However, when the number of trainable parameters is far more than that of training data, deep networks may suffer from overfitting. This leads to the routine usage of regularization methods such as data augmentation \cite{autoaugment}, weight decay \cite{cifar}, Dropout \cite{dropout} and Batch Normalization~(BN) \cite{bn} to prevent overfitting and improve generalization.

Although regularization has been an essential part of deep learning, deciding which regularization methods to use remains an art. Even if each of the regularization methods works well on its own, combining them does not always give improved performance. For instance, the network trained with both Dropout and BN may not produce a better result \cite{bn, BNandDropout}. Dropout may change the statistical variance of layers output when we switch from training to testing, while BN requires the variance to be the same during both training and testing~\cite{BNandDropout}.

\begin{figure}%
    \centering
    \subfloat[Standard neural network with nonlinearity.]{{\includegraphics[width=0.3\textwidth]{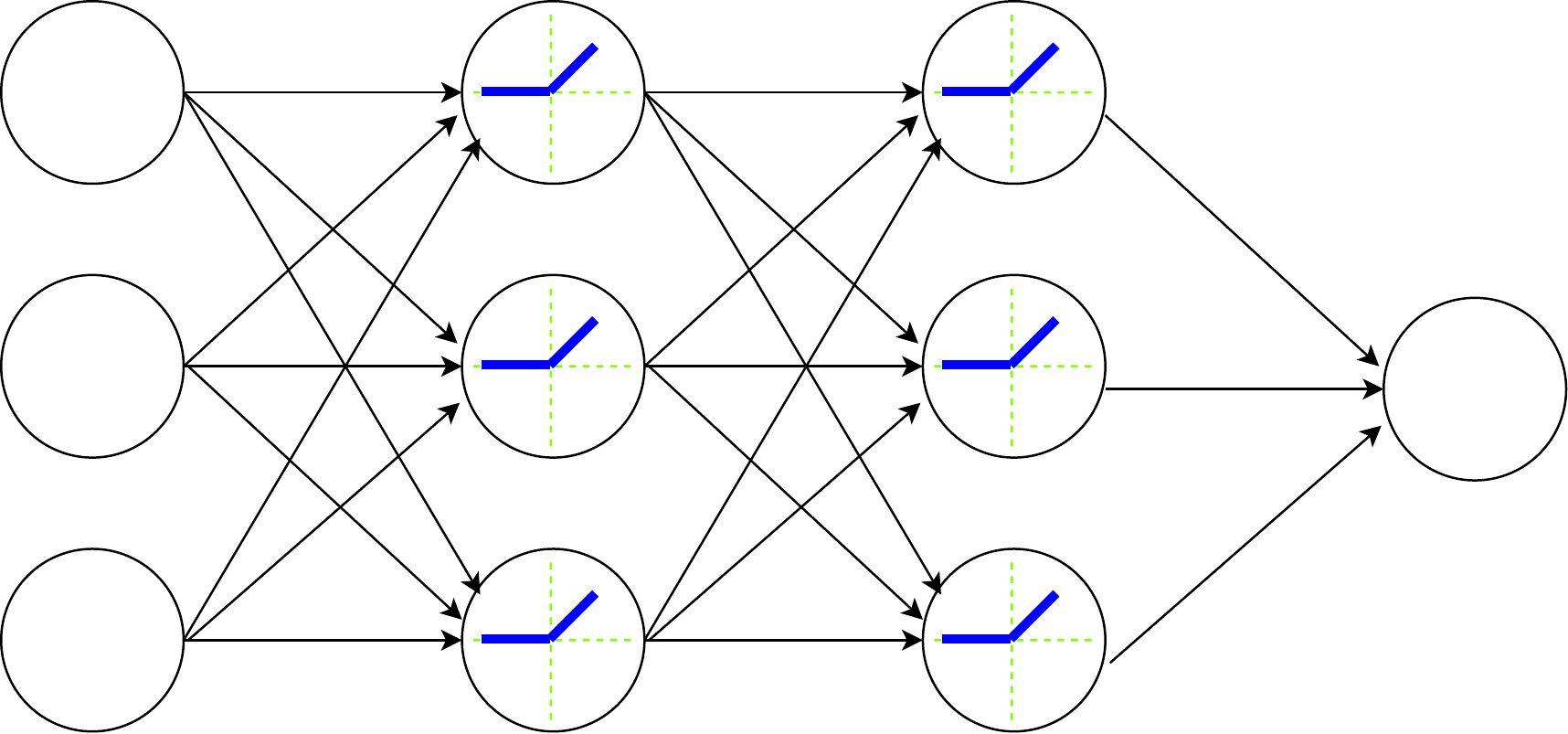} }\label{fig:std-act}}%
    \qquad
    \qquad
    \subfloat[After applying Drop-Activation during training.]{{\includegraphics[width=0.3\textwidth]{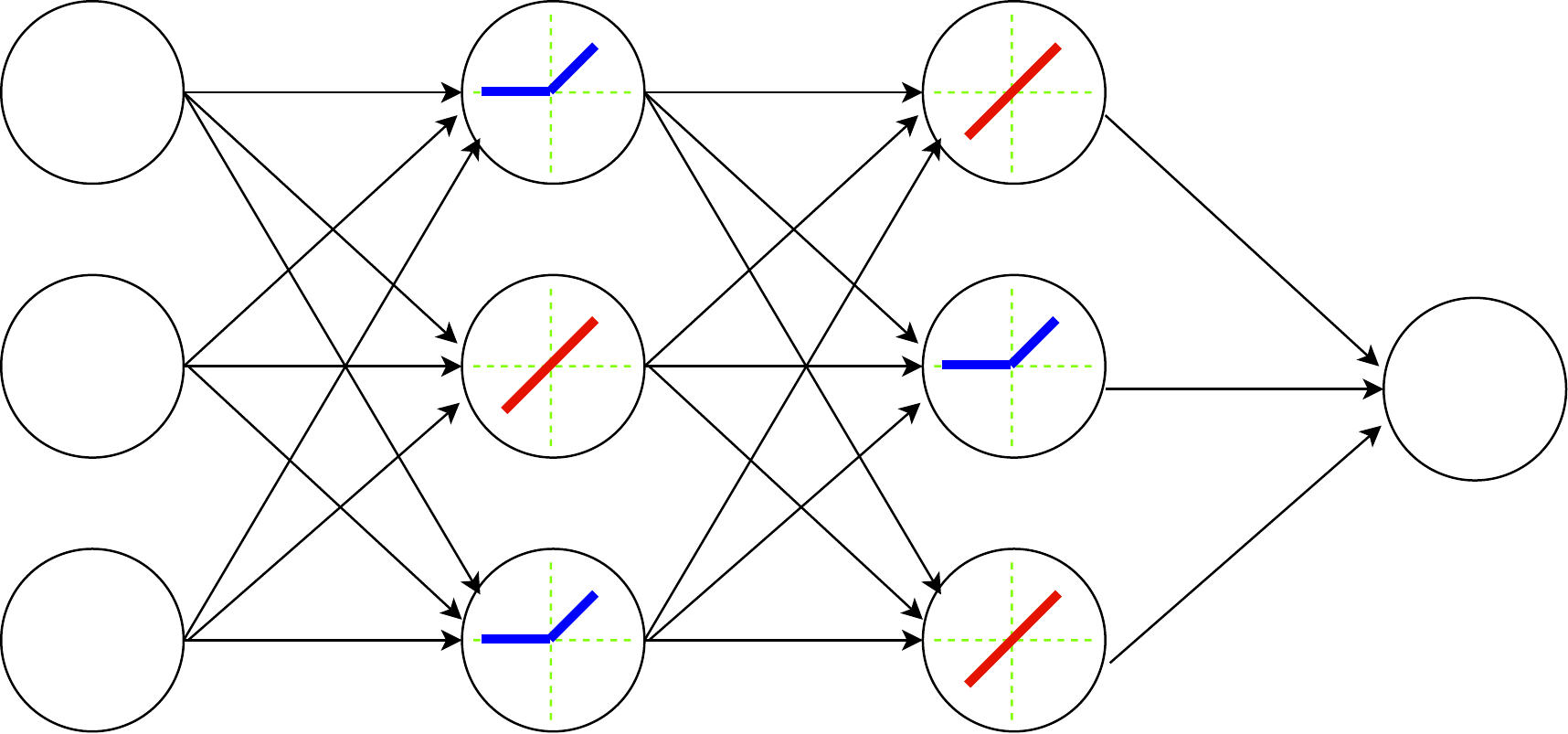}}\label{fig:drop-act}}%
    \caption{Illustration of Drop-Activation. \textbf{Left:} A standard 2-hidden-layer network with nonlinear activation (Blue). \textbf{Right:} A new network generated by applying Drop-Activation to the network on the left. Nonlinear activation functions are randomly selected and replaced with identity maps (Red).}%
    \label{fig:DropActMethod}%
    \vspace{-0.5cm}
\end{figure}

\textbf{Our contributions:} To deal with the aforementioned challenges, we propose a novel regularization method, Drop-Activation, inspired by the works in \cite{dropout, cutout, stochasticdepth, shakedrop, dropconnect, swapout, zoneout}, where some structures of networks are dropped to achieve better generalization. The advantages are as follows:
\begin{itemize}
    \item Drop-Activation provides an easy-to-implement yet effective method for regularization via implicit parameter reduction.
    \item Drop-Activation can be used in synergy with the most popular architectures and regularization methods, leading to improved performance in various datasets for image classification.
\end{itemize}
The basic idea of Drop-Activation is that the nonlinearities in the network will be randomly activated or deactivated during training. More precisely, the nonlinear activations are turned into identity mappings with a certain probability, as shown in Figure~\ref{fig:DropActMethod}. At testing time, we propose using a deterministic neural network with a new activation function which is a combination of identity mapping and the dropped nonlinearity, to represent the ensemble average of the random networks generated by Drop-Activation. Rectified linear unit~(ReLU) has the advantage of reducing the saturation of gradient and accelerating the training compared with sigmoid or tanh activation function~\cite{verydeep}. It is frequently adopted in modern deep neural networks~\cite{resnet, preresnet, densenet, wrn, resnext}. In this paper, we focus on studying the random replacement of the ReLU activation function with the identity function.

The starting point of Drop-Activation is to randomly draw an ensemble of neural networks with either an identity mapping or a ReLU activation function. The training process of Drop-Activation is to identify a set of parameters such that various neural networks in this ensemble work well when being assigned with these parameters. By ``fitting'' to many neural-networks instead of a fixed one, overfitting can potentially be prevented. Indeed, our theoretical analysis shows that Drop-Activation implicitly adds a penalty term to the loss function, aiming at network parameters such that the corresponding deep neural network can be approximated by a linear network, \ie, implicit parameter reduction.

\textbf{Organizations:} The remainder of this paper is structured as follows. In Section \ref{sec:2}, we review some of the regularization methods and discuss their relations with our work. In Section \ref{sec:3}, we formally introduce Drop-Activation. In Section \ref{sec:5}, the theoretical analysis demonstrates the regularization of Drop-Activation and its synergy with BN. In Section \ref{sec:4}, these advantages of Drop-Activation are further supported by our numerical experiments carried on different datasets and networks.

\section{Related Work}
\label{sec:2}
Various regularization methods have been proposed to reduce the risk of overfitting. Data augmentation achieves regularization by directly enlarging the original training dataset via randomly transforming the input images \cite{verydeep, simonyan2014very, cutout, autoaugment} or output labels \cite{mixup, disturblabel}. Another class of methods regularize the network by adding randomness into various neural network structures such as nodes \cite{dropout}, connections \cite{dropconnect}, pooling layers \cite{pool}, activations \cite{rrelu} and residual blocks \cite{shake, stochasticdepth, shakedrop}. In particular \cite{dropout, cutout, stochasticdepth, shakedrop, dropconnect, swapout, zoneout} add randomness by dropping some structures of neural networks at random in training.

Dropout \cite{dropout} drops nodes along with its connection with some fixed probability during training. DropConnect \cite{dropconnect} has a similar idea but masks out some weights randomly. \cite{stochasticdepth} improves the performance of ResNet \cite{resnet} by dropping the entire residual block at random during training and passing through skip connections (identity mapping). This idea is also used in \cite{shakedrop} when training ResNeXt \cite{resnext} type 2-residual-branch network. The idea of dropping also arises in data augmentation. Cutout \cite{cutout} randomly cuts out a square region of training images to prevent the neural network from putting too much emphasis on the specific region of features.

A related idea of dropping ``functions'' in neural networks were proposed in~\cite{swapout, zoneout}, where subnetwork structures are discarded randomly, instead of replacing an activation function with the identify function. \cite{swapout} proposes a framework of Swapout $\Theta_1 \otimes X + \Theta_2\otimes F(X)$, where $X$ is the input feature map, $F$ is a sub-network, $\Theta_1$ and $\Theta_2$ are i.i.d Bernoulli distribution, and $\otimes$ is the element-wise product. \cite{swapout} verified the effectiveness of the proposed Swapout framework on a large structure, i.e., $F$ is a residual in ResNet~\cite{resnet} which consists of layers of BN, ReLU, Convolution. Similarly, Zoneout~\cite{zoneout} discussed the case when $F$ is a set of layers. Dropping a subnetwork structure can lead to instability of training and it requires more careful hyperparameter tuning. On the contrary, Drop-Activation focuses on the nonlinear activation functions, a smaller and more basic structure of networks.We will show the effectiveness of Drop-Activation on a wider range of networks and datasets than those in~\cite{swapout, zoneout}.

In the next section, inspired by the above methods, we propose the Drop-Activation method for regularization. We want to emphasize that the improvement by Drop-Activation is universal to most neural-network architectures, and it can be readily used in conjunction with many regularizers without conflicts.

\section{Formulation of Drop-Activation}
\label{sec:3}
This section describes the Drop-Activation method. Suppose $x_0$ is an input vector of an $L$-layer feed forward network. Let $x_l$ be the output of $l$-th layer. $f(\cdot)$ is the element-wise nonlinear activation operator that maps an input vector to an output vector by applying a nonlinearity on each of the entries of the input. Without the loss of generality, we assume $f:\mathbb{R}^d\rightarrow \mathbb{R}^d$, \eg,
% \begin{equation}
% f(x) = \begin{bmatrix} \sigma\left(x[1]\right) \\ \vdots \\ \sigma\left(x[d]\right) \end{bmatrix}\in \mathbb{R}^d, \quad x = \begin{bmatrix} x[1] \\ \vdots \\ x[d] \end{bmatrix}\in \mathbb{R}^d,
% \end{equation}
\begin{equation}
f(x) = \begin{bmatrix} \sigma\left(x[1]\right), \cdots,  \sigma\left(x[d]\right) \end{bmatrix}^T\in \mathbb{R}^d, \quad x = \begin{bmatrix} x[1] , \cdots, x[d] \end{bmatrix}^T\in \mathbb{R}^d,
\end{equation}
where $\sigma$ could be a ReLU, a sigmoid or a tanh function but we only consider that $\sigma$ is a ReLU function in our paper. For a standard fully connected or convolution network, the $d$-dimensional output can be written as
\begin{align}
x_{l+1} = f(W_lx_l),
\label{eq:output-L}
\end{align}
where $W_l\in \mathbb{R}^{d\times d}$ is the weight matrix of the $l$-th layer. Biases are neglected for the convenience of presentation.

In what follows, we modify the way of applying the nonlinear activation operator $f$ to achieve regularization.
In the training phase, we remove the pointwise nonlinearities in $f$ randomly. In the testing phase, the function $f$ is replaced with a new deterministic nonlinearity.

\textbf{Training Phase:}  During training, the $d$ nonlinearities $\sigma$ in the operator $f$ are kept with probability $p$ (or dropping them with probability $1-p$). The output of the $(l+1)$-th layer is thus
\begin{align}
\begin{split}
x_{l+1} &= (I-P)W_lx_l + Pf(W_lx_l) = (I-P+Pf)(W_lx_l),
\end{split}
\label{eq:output-L-da-train}
\end{align}
where $P = \text{diag}(P_1,P_2,\cdots,P_d)$, $P_1,\cdots,P_d$ are independent and identical random variables following a Bernoulli distribution $B(p)$ that takes value $1$ with probability $p$ and $0$ with probability $1-p$. We use $I$ to denote the identity matrix. Intuitively, when $P = I$, then $x_{l+1} = f(W_lx_l)$, meaing all the nonlinearities in this layer are kept. When $P = \mathbf{0}$ , then $x_{l+1} = W_lx_l$, meaning all the nonlinearities are dropped. The general case lies somewhere between these two limits where the nonlinearities are kept or dropped partially. At each iteration, a different realization of $P$ is sampled from the Bernoulli distribution again. %The nonlinear activation for a specific node have chance to become identity. Therefore, we train a fixed architecture  but with different activation operators at each iteration.

When the nonlinear activation function in Eqn.~\eqref{eq:output-L-da-train} is ReLU, the $j$-th component of $(I-P+Pf)(x)$ can be written as
\begin{equation}
(I-P+Pf)(x)[j] = \left\{
\begin{aligned}
x[j]&, \quad x[j]\geq 0,\\
(1-P_j)x[j]&, \quad x[j]<0.
\end{aligned}
\right.
\label{eq:droprelu}
\end{equation}

\textbf{Testing Phase:} During testing, we use a deterministic nonlinear function resulting from averaging the realizations of $P$. More precisely, we take the expectation of the Eqn.~\eqref{eq:output-L-da-train} with respect to the random variable $P$:
\begin{align}
\begin{split}
x_{l+1} &= \mathbb{E}_{P_i\sim \ B(p)} (I-P+Pf)(W_lx_l) = ((1-p)I + pf)(W_lx_l),
\end{split}
\label{eq:output-L-da-test}
\end{align}
and the new activation function $(1-p)I + pf$ is the convex combination of an  identity operator $I$ and an activation operator $f$. Eqn.~\eqref{eq:droprelu} is the deteministic nonlinearity used to generate a deterministic neural network for testing. In particular, when ReLU is used, then the new activation $(1-p)I + pf$ is the leaky ReLU with slope $1 - p$ in its negative part~\cite{rrelu}.

\section{Theoretical Analysis}
\label{sec:5}
In Section \ref{subsec:regularizer}, we show that in a ReLU neural-network with one-hidden-layer, Drop-Activation provides a regularization via penalizing the difference between nonlinear activation network and linear network, which can be understood as implicit parameter reduction, \ie, the intrinsic dimension of the parameter space is reduced. In Section \ref{subsec:harmony}, we further show that the use of Drop-Activation does not impact some other techniques such as BN, which ensures the practicality of using Drop-Activation in deep networks.

\subsection{Drop-Activation as a regularizer}
\label{subsec:regularizer}
We use similar ideas in \cite{dropout} and \cite{dropanalysis} to show that having Drop-Activation in a standard one-hidden layer fully connected neural network with ReLU activation gives rise to an explicit regularizer.

Let $x$ be the input vector, $y$ be the output. The output of the one-hidden layer neural ReLU network is $\hat{y} = W_2r(W_1x)$, where $W_1$, $W_2$ are weights of the network, $r:\mathbb{R}^d\rightarrow \mathbb{R}^d$ is the function for applying ReLU elementwise to the input vector. Let $r_p(\cdot)$ denotes the leaky ReLU with slope $1-p$ in the negative part. As in Eqn.~\eqref{eq:output-L-da-train} and \eqref{eq:output-L-da-test}, applying Drop-Activation to this network gives
\begin{align}
\hat{y} = W_2((I-P+Pr)W_1x)
\label{eq:output-train}
\end{align}
during training, and
\begin{align}
\hat{y} = W_2((1-p)I+pr)W_1x = W_2r_p(W_1x)
\label{eq:output-test}
\end{align}
during testing. Suppose we have $n$ training samples $\{(x_i,y_i)\}_{i=1}^n$. To reveal the effect of Drop-Activation, we average the training loss function over $P$:
\begin{align}
\begin{split}
\min\limits_{W_1,W_2}& \sum_{i=1}^n \mathbb{E} \|W_2[(I-P+Pr)W_1x_i] - y_i\|_2^2,
\end{split}
\label{opt:train}
\end{align}
where the expectation is taken with respect to the feature noise $P_1,\cdots,P_d$. The use of Drop-Activation can be seen as applying a stochastic minimization to such an average loss. The result after averaging the loss function over $P$ is summarized as follows.
\begin{property}
	The optimization problem (\ref{opt:train}) is equivalent to
	\begin{align}
	\begin{split}
	\min\limits_{W_1,W_2}\sum_{i=1}^n \|W_2r_p(W_1x_i) - y_i\|_2^2 + p^{-1}(1-p)\|W_2W_1x_i - W_2r_p(W_1x_i)\|_2^2.
	\end{split}
	\label{opt:3}
	\end{align}
	\label{prop1}
	\vspace{-0.5cm}
\end{property}

\begin{proof}
Suppose that $x$ is the input vector. Let $D_{W_1,x} = \text{diag}\{(W_1x>0)\}$, where $(W_1x>0)$ is a 0-1 vector, and the j-th component of $(W_1x>0)$ is equal to 1 if the j-th component of $W_1x$ is positive or is equal to 0 else. Then, the ReLU mapping of $W_1x$ can be written as $r(W_1x)=D_{W_1,x}W_1x$. For simplification, we denote
\begin{align*}
S := I-P+PD_{W_1,x}, \ \ S_p:= I-pI+pD_{W_1,x}, \ \ v := W_1x.
\end{align*}

On one hand, $\|W_2r_p(W_1x) - y\|_2^2 = \|W_2S_pv-y\|_2^2$. We expand it and obtain
\begin{align}
\begin{split}
\|W_2S_pW_1x-y\|_2^2 = \text{tr}(W_2S_pvv^TS_pW_2^T)-2\text{tr}(W_2S_pvy^T)+\text{tr}(yy^T),
\end{split}
\label{eq:avg}
\end{align}
where function $\text{tr}(\cdot)$ is the trace operator computing the sum of matrix diagonal. We denote $\text{vec}(\cdot)$ as a function converting the diagonal matrix into a column vector. Rewrite the first term of Eqn.~\eqref{eq:avg} and get
\begin{align}
\begin{split}
\text{tr}(W_2S_pvv^TS_pW_2^T) = &\text{tr}(S_pvv^TS_pW_2^TW_2)
\\ = &\text{tr}(\text{diag}(v)\text{vec}(S_p)\text{vec}(S_p)^T\text{diag}(v)W_2^TW_2)
\\ = &\text{tr}(\text{vec}(S_p)\text{vec}(S_p)^T\text{diag}(v)W_2^TW_2\text{diag}(v)).
\end{split}
\label{eq:Sp}
\end{align}

On the other hand, we have
\begin{align}
\begin{split}
\mathbb{E}\|W_2[(I-P+Pr)W_1x] - y\|_2^2 =& \mathbb{E}[\|W_2Sv-y\|_2^2]
\\ =& \mathbb{E}[\text{tr}(W_2Svv^TSW_2^T)]-2\text{tr}(W_2S_pvy^T)+\text{tr}(yy^T),
\end{split}
\label{eq:not_avg}
\end{align}
where the expectation is taken with respect to the feature noise $P = \{P_1,\cdots,P_d\}.$ Similar to Eqn.~\eqref{eq:Sp}, we combine the matrices containing random variables and obtain
\begin{align}
\begin{split}
\text{tr}(W_2Svv^TSW_2^T) = \text{tr}(\text{vec}(S)\text{vec}(S)^T\text{diag}(v)W_2^TW_2\text{diag}(v)).
\end{split}
\label{eq:S}
\end{align}

Since $\text{tr}(\cdot)$ has property of linearity, taking the expectation of Eqn. (\ref{eq:S}) with respect to $P$ obtains
\begin{align}
\begin{split}
\mathbb{E}\text{tr}(W_2Svv^TSW_2^T)=\text{tr}(\mathbb{E}(\text{vec}(S)\text{vec}(S)^T)\text{diag}(v)W_2^TW_2\text{diag}(v)).
\end{split}
\label{eq:S_part}
\end{align}

Denote $D_{W_1,x}= \text{diag}(d_1,\cdots,d_k)$, and then
\begin{align}
\begin{split}
\mathbb{E}[\text{vec}(S)\text{vec}(S)^T]-\text{vec}(S_p)\text{vec}(S_p)^T
 =& \text{diag}(\{\mathbb{E}((1-P_i+P_id_i)^2)-(1-p+pd_i)^2\}_{i=1}^k)
\\ =& p(1-p)(I-D_{W_1,x})^2.
\end{split}
\label{eq:S_simp}
\end{align}

Using Eqn.~\eqref{eq:S_simp}, Eqn.~\eqref{eq:Sp} and Eqn.~\eqref{eq:S}, we can get the difference between Eqn.~\eqref{eq:avg} and Eqn.~\eqref{eq:not_avg},
\begin{align*}
&  \mathbb{\mathbb{E}}[\text{tr}(W_2Svv^TSW_2^T)] - \text{tr}(W_2S_pvv^TS_pW_2^T)
\\ =& \text{tr}\{(\mathbb{E}(\text{vec}(S)\text{vec}(S)^T)-\text{vec}(S_p)\text{vec}(S_p)^T)\text{diag}(v)W_2^TW_2\text{diag}(v)\}
\\ =& p(1-p)tr\{(I-D_{W_1,x})^2\text{diag}(v)W_2^TW_2\text{diag}(v)\}
\\ =& p(1-p)tr\{W_2\text{diag}(v)(I-D_{W_1,x})^2\text{diag}(v)W_2^T\}
\\ =& p(1-p)\|W_2(I-D_{W_2,x})W_1x\|_2^2.
\end{align*}

Note that $D_{W_1,x} - I = \frac{1}{p}(S_p - I)$, so we have
\begin{align*}
p(1-p)\|W_2(I-D_{A,x})W_1x\|_2^2
= \frac{1-p}{p} \|W_2(I-S_p)W_1x\|_2^2
=  \frac{1-p}{p} \|W_2W_1x-W_2r_p(W_1x)\|_2^2.
\end{align*}

Finally, we attain the difference between Eqn.~\eqref{eq:avg} and Eqn.~\eqref{eq:not_avg},
$$\frac{1-p}{p} \|W_2W_1x-W_2r_p(W_1x)\|_2^2.$$
\end{proof}

We refer to the objective function of the optimization~(\ref{opt:3}). The first term is nothing but the $l_2$ loss during prediction time $\sum_i \| \hat y_i - y_i \|_2^2$, where $\hat y_i$'s are defined via \eqref{eq:output-test}. Therefore, Property \ref{prop1} shows that Drop-Activation incurs a penalty
\begin{equation}
p^{-1}(1-p)\|W_2W_1x_i - W_2r_p(W_1x_i)\|_2^2
\label{eq:penalty}
\end{equation}
on top of the prediction loss. In Eqn. (\ref{eq:penalty}), the coefficient $\frac{1-p}{p}$ influences the magnitude of the penalty. In our experiments, $p$ is selected to be a large number close to $1$ (typically $0.95$). The magnitude of the penalty will not be large in our numerical experiments.

%Compared Eqn.(\ref{opt:test}) with Eqn.(\ref{opt:3}) the penalty term is formulated as
%\begin{align}
%\frac{1-p}{p}\|W_2W_1x_i - W_2r_p(W_1x_i)\|_2^2,
%\label{eq:penalty}
%\end{align}
%which is composed of coefficient $\frac{1-p}{p}$ and $l_2$ penalty.

%The difference between the objective functions in optimization \eqref{opt:test} and \eqref{opt:train} acts as a regularizer that incorporates the effect of the randomness from dropping the activation operator. We extract the regularizer in Property \ref{prop1}.

The penalty (\ref{eq:penalty}) consists of the terms $W_2W_1x$ and $W_2r_p(W_1x)$. $W_2W_1x$ has no nonlinearity, so it is a linear network. In contrast, since $W_2r_p(W_1x)$ has the nonlinearity $r_p$,  it can be considered as a deep network. The two networks share the same parameters $W_1$ and $W_2$. Therefore the penalty (\ref{eq:penalty}) encourages weights $W_1,W_2$ such that the prediction of the relatively deep network $W_2r_p(W_1x)$ should be somewhat close to that of a linear network. %\sout{In a classification or regression task, the shallow network has less representation power but the lower parameter complexity of the shallow network results in mappings with better generalization property.}
In this way, the penalty incurs by Drop-Activation may help in reducing overfitting by implicit parameter reduction. %{\color{red}In a classification or regression task, the shallow network has less representation power but the lower parameter complexity of the shallow network results in mappings with better generalization property. In this way, the penalty incurs by Drop-Activation may help in reducing overfitting by implicit parameter reduction.} {\color{green} when changed to linear network, this statement has problem.}

\begin{figure*}%
    \centering
    \subfloat[The ground true function: $x\sin x$.]{{\includegraphics[width=0.37\textwidth]{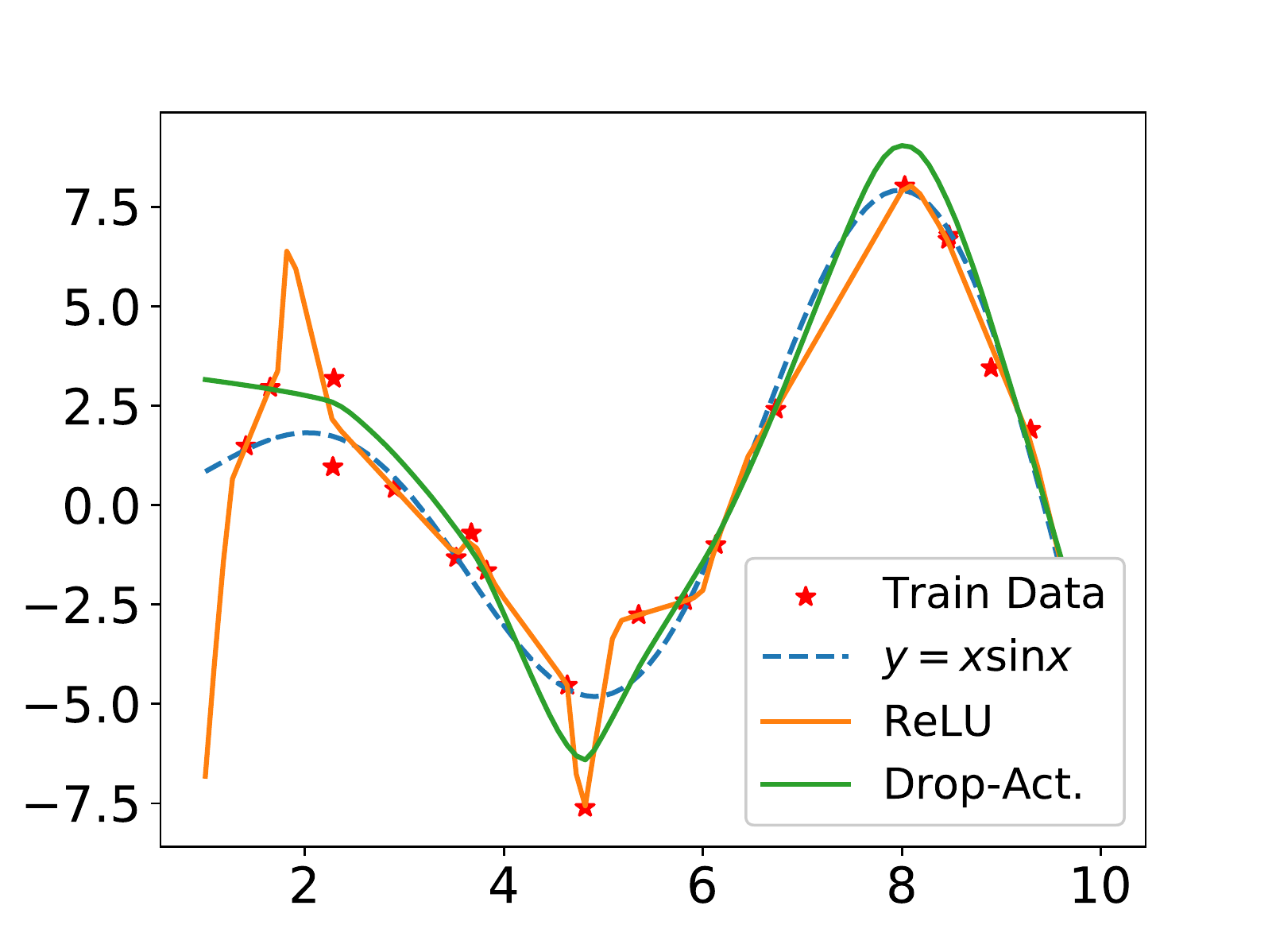} }\label{fig:r1}}%
    \qquad
    \subfloat[The ground true function: A piecewise constant function.]{{\includegraphics[width=0.37\textwidth]{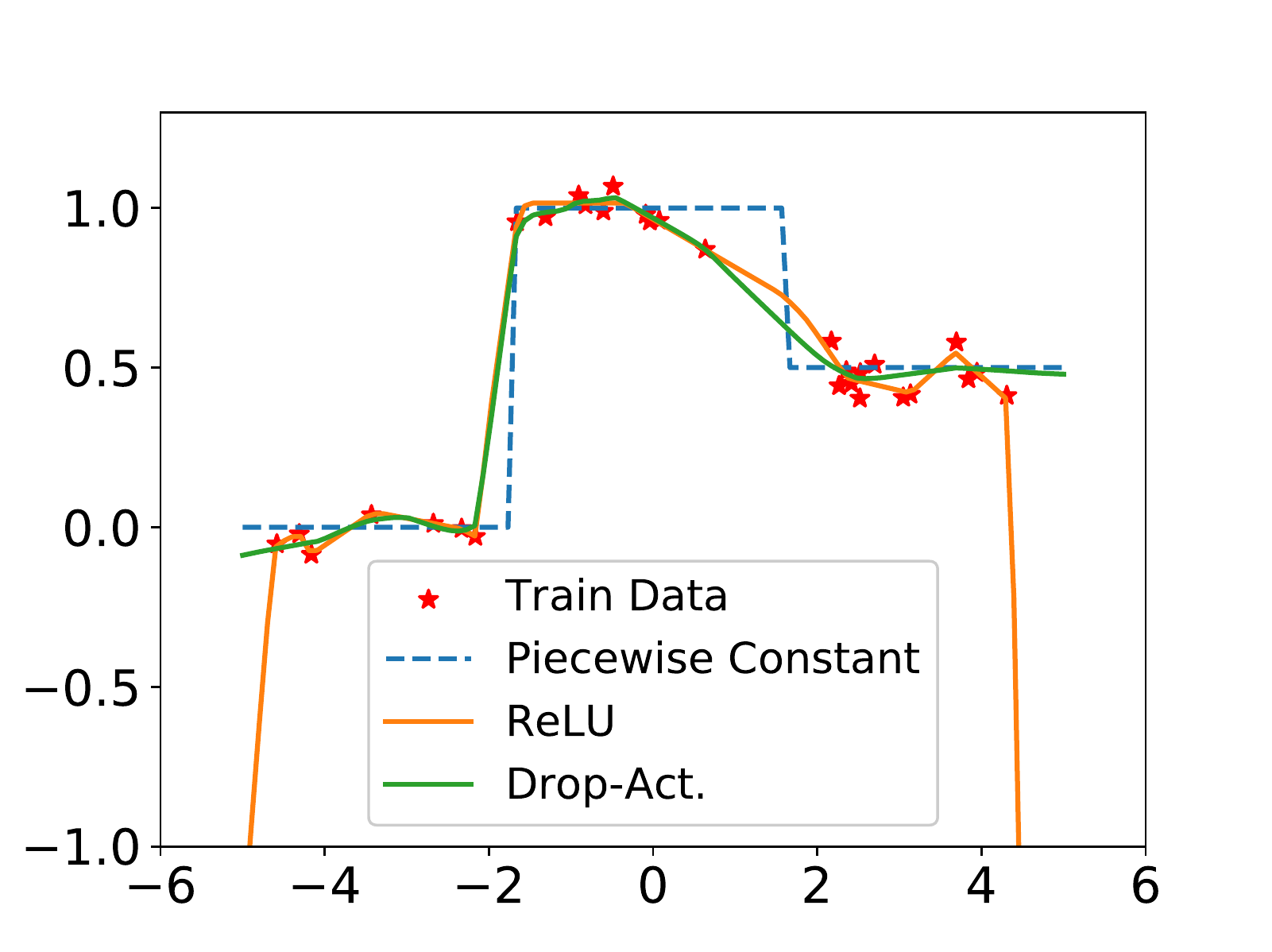}}\label{fig:r2}}%
    \caption{Comparison between the networks equipped with Drop-Activation and normal ReLU. (a) Regression of $x\sin x$. (b) Regression of a piecewise constant function. Blue:  Ground truth functions. Orange: Regression results using ReLU. Green: Regression results using Drop-Activation. ``$*$'': Training data perturbed by Gaussian noise.}%
    \label{fig:reg}%
    \vspace{-0.5cm}
\end{figure*}

To illustrate this point, we perform a simple regression task for two functions. To generate the training dataset, we sample 20 $(x_i, y_i)$ pairs from the ground truth function and add gaussian noise on the outputs. Then we train a fully connected network with three hidden layers of width 1000, 800, 200, respectively. Figure \ref{fig:r1} and \ref{fig:r2} show that the network with ReLU has a low prediction error on training data points, but is generally erroneous in other regions. Although the network with Drop-Activation does not fit as well to the training data (comparing with using normal ReLU), overall it achieves a lower prediction error. With the effect of incurred penalty \eqref{eq:penalty}, the network with Drop-Activation reduces the influence of data noise and yields a smooth curve.

Figure~\ref{fig:resnet-164} shows the training of ResNet164 on CIFAR100, the training error with Drop-Activation is slightly larger than that without Drop-Activation. However, in terms of the generalization error, Drop-Activation gives improved performance. This verifies that the original network has been over-parametired and Drop-Activation can regularize the network by implicit parameter reduction.

\subsection{Compatibility of Drop-Activation with BN}
\label{subsec:harmony}
In this section, we show theoretically that Drop-Activation essentially keeps the statistical property of the output of each network layer when going from training to testing phase and hence it can be used together with BN. \cite{BNandDropout} argues that BN assumes the output of each layer has the same variance during training and testing. However, Dropout~\cite{dropout} will shift the variance of the output during the testing time leading to disharmony when used in conjunction with BN. Using a similar analysis as \cite{BNandDropout}, we show that unlike Dropout, Drop-Activation can be used together with BN since it maintains the output variance.

\begin{figure}[ht]
\centering
%\subfloat[]{
\begin{minipage}[t]{0.4\textwidth}
\vspace{0pt}
\includegraphics[width=0.95\textwidth]{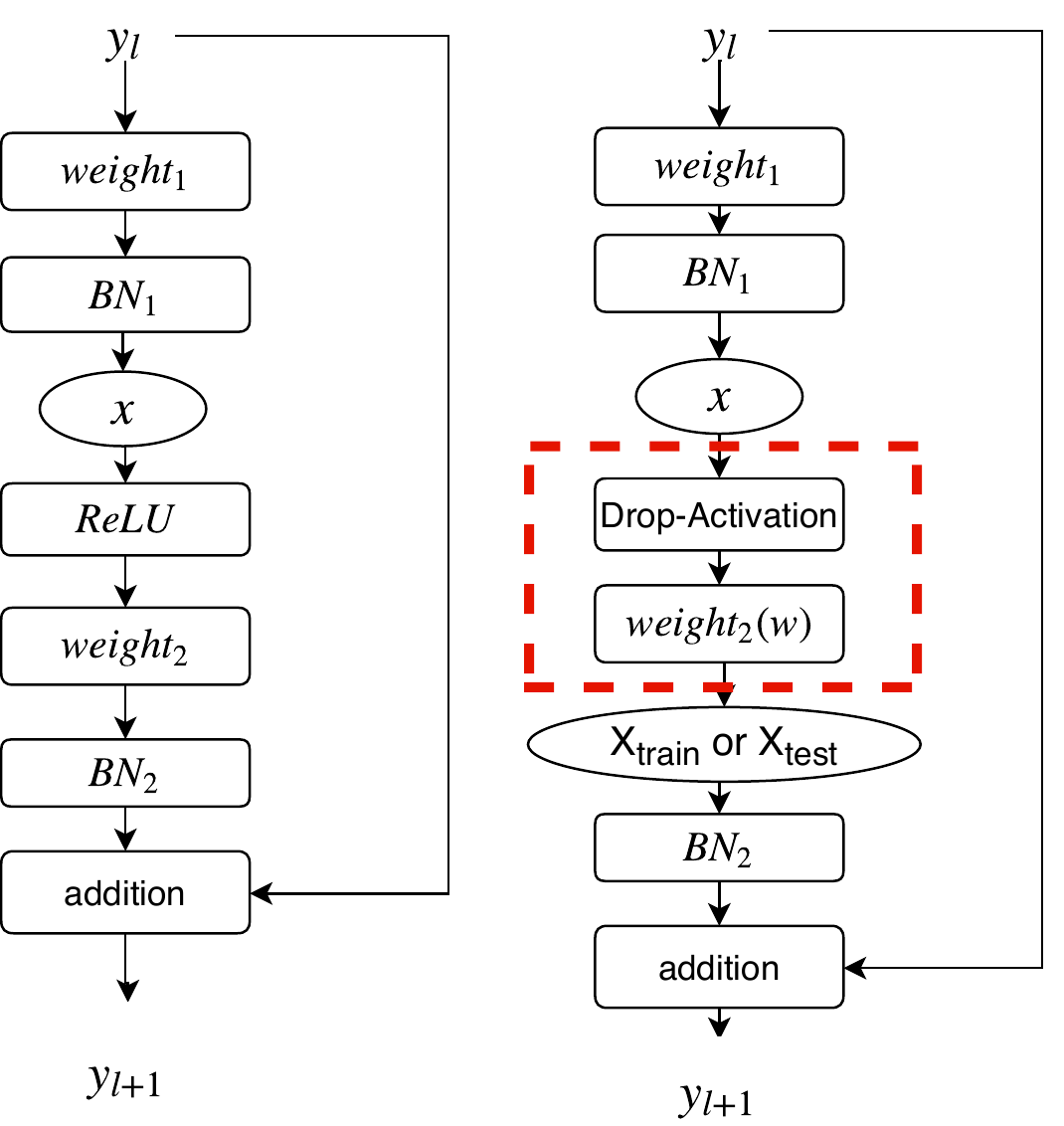}
\caption{\textbf{Left}: A basic block in ResNet. \textbf{Right}: A basic block of a network with Drop-Activation. }
\label{fig:basicblock}
\end{minipage}
\hspace{.3in}
%\subfloat[]{
\begin{minipage}[t]{0.4\textwidth}
\vspace{0pt}
\includegraphics[width=1\textwidth]{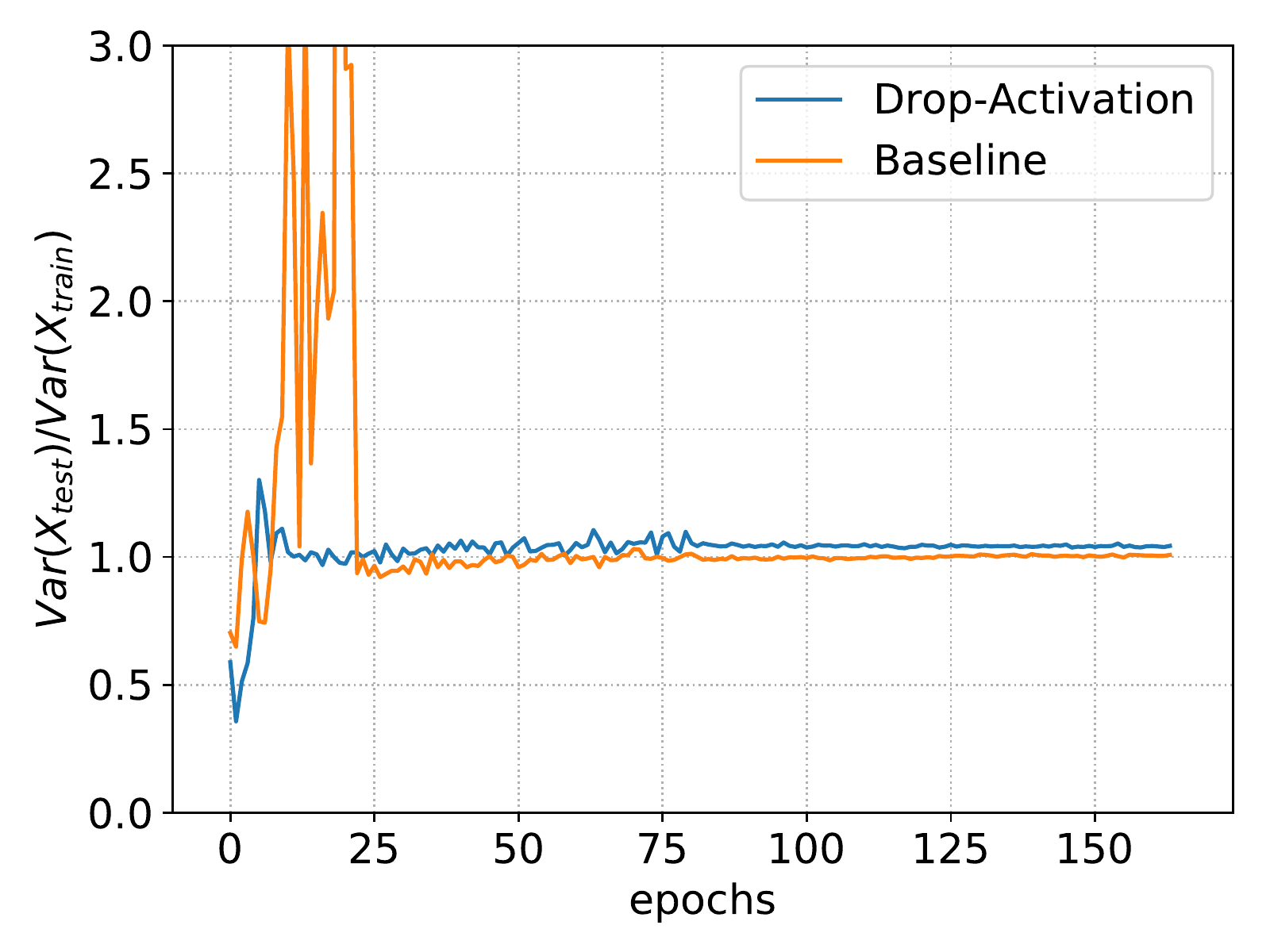}
\caption{The shift ratio of the output of the second stage for ResNet-164.
  $\text{Var}(X_\text{train})$ and $\text{Var}(X_\text{test})$ denote the average of the variance for the output of the second stage during training and testing respectively.}
\label{fig:var_shift}
\end{minipage}
\vspace{-0.4cm}
\end{figure}

To this end, we analyze the mappings in ResNet \cite{resnet}. Figure \ref{fig:basicblock} (Left) shows a basic block of ResNet while Figure \ref{fig:basicblock} (Right) shows a basic block with Drop-Activation. We focus on the rectangular box with dashed line. Suppose the output from the $BN_1$ shown in Figure \ref{fig:basicblock} is $x = (x[1],\cdots,x[d])$. \cite{lee2019wide} shows the hidden features converge in distribution to the Gaussian when $d$ is large, so for simplification, we assume that $x[i]\sim\ \mathcal{N}(0,1),\ i=1,\ldots,d$ are i.i.d. random variables. When $x$ is passed to the Drop-Activation layer followed by a linear transformation $weight_2$ with weights $w= (w_1,\cdots,w_d) \in \mathbb{R}^{1\times d} $, we obtain
$
X_\text{train}:=\sum_{i=1}^d w_{i}((1-P_i)x[i] + P_ir(x[i])),
$
where $P = \text{diag}(P_1,\cdots,P_d)$ and $P_i\sim\ B(p)$. Similarly, during testing, taking the expectation over $P_i$'s gives
$
X_\text{test}:=\sum_{i=1}^d w_{i}((1-p)x[i] + pr(x[i])).
$
The output of the rectangular box $X_\text{train}$ (and $X_\text{test}$ during testing) is then used as the input to $BN_2$ in Figure~\ref{fig:basicblock}. Since for BN we only need to understand the entry-wise statistics of its input, without loss of generality, we assume the linear transformation $w$ maps a vector from $\mathbb{R}^d$ to $\mathbb{R}$, $X_\text{train}$ and $X_\text{test}$ are scalars.

We want to show $X_\text{train}$ and $X_\text{test}$ have similar statistics. By design, $\mathbb{E}_{P,x} X_\text{train}= \mathbb{E}_{P,x} X_\text{test}$. Notice that the expectation here is taken with respect to both the random variables $P$ and the input $x$ of the box in Figure~\ref{fig:basicblock}. Thus the main question is whether the variances of $X_\text{train}$ and $X_\text{test}$ are the same. To this end, we introduce the Shift Ratio \cite{BNandDropout}:
% $
% \text{Shift ratio} = \frac{\text{Var}(X_\text{test})}{\text{Var}(X_\text{train})}
% $
$
\text{Shift Ratio} = \text{Var}(X_\text{test})/\text{Var}(X_\text{train})
$
 as a metric for evaluating the variance shift. The shift ratio is expected to be close to $1$, since the BN layer $BN_2$ requires its input having similar variance in both training and testing time.
\begin{property}
The shift ratio of $X_\text{train}$ and $X_\text{test}$ is
\begin{align}
\begin{split}
\text{Var}(X_\text{test})/\text{Var}(X_\text{train}) = [(\pi -1)p^2 - 2\pi p + 2\pi]/[2\pi - \pi p -p^2].
\end{split}
\label{formula}
\end{align}
\label{prop2}
\vspace{-0.5cm}
\end{property}

\begin{proof}
Since $x[i]\sim \mathcal{N}(0,1),$ it is easy to get
$\mathbb{E}(x[i]) = 0$, $\mathbb{E}(r(x[i])) = \frac{1}{\sqrt{2\pi}}$, $\mathbb{E}(x[i]^2) = 1$, and $\mathbb{E}(r(x[i])^2) = \frac{1}{2}$,
where the expectation is taken with respect to random variable $x[i].$
We have
\begin{align*}
&\mathbb{E}(X_\text{train})= \sum_{i=1}^dw_i\mathbb{E}((1-P_i+P_ir)x[i])= \frac{p\sum_{i=1}^dw_i}{\sqrt{2\pi}},\\
&\mathbb{E}(X_\text{test})=  \sum_{i=1}^dw_i\mathbb{E}((1-p+pr)x[i])= \frac{p\sum_{i=1}^dw_i}{\sqrt{2\pi}},
\end{align*}
 where expectation is taken with respect to feature noise $P=\{P_1,\cdots,P_d\}$ and inputs $(x[1],\cdots,x[d])$. In what follows, we compute $\text{Var}(X_\text{train})$ and $\text{Var}(X_\text{test})$.

Expand the square of $X_\text{train}$ to get
\begin{align*}
X_\text{train}^2 = \sum_{i=1}^d w_i^2((1-P_i)x[i] + P_ir(x[i]))^2 + 2\sum_{i< j}w_iw_j((1-P_i)x[i] + P_ir(x[i]))((1-P_j)x[j] + P_jr(x[j])).
\end{align*}

Then we obtain its expectation,
\begin{align*}
\mathbb{E}(X_\text{train}^2) &= \sum_{i=1}^d w_i^2\mathbb{E}((1-P_i)^2x[i]^2 + 2(1-P_i)P_ix[i]r(x[i]) + P_i^2r(x[i])^2) + 2\sum_{i< j}w_iw_j \mathbb{E}(P_iP_jr(x[i])r(x[j]))
\\ & = \sum_{i=1}^d w_i^2(1-p + \frac{1}{2}p) +  \frac{p^2}{\pi}\sum_{i< j}w_iw_j.
\end{align*}

Using the fact that $\text{Var}(X_\text{train}) = \mathbb{E}(X_\text{train}^2) - (\mathbb{E}X_\text{train})^2$, we get
\begin{align}
\begin{split}
\text{Var}(X_\text{train})
= \sum_{i=1}^d w_i^2(1-p + \frac{1}{2}p) +  \frac{p^2}{\pi}\sum_{i< j}w_iw_j - (\frac{1}{\sqrt{2\pi}}p\sum_{i=1}^dw_i)^2
 = \sum_{i=1}^d w_i^2(1-\frac{1}{2}p - \frac{1}{2\pi}p^2).
\end{split}
\label{eq:xtrain_var}
\end{align}

So far, we have finished $\text{Var}(X_\text{train})$. Now we are going to compute $\text{Var}(X_\text{test})$. Expand $X_\text{test}^2$ to get
\begin{align*}
X_\text{test}^2 = \sum_{i=1}^d w_i^2((1-p)x[i] + pr(x[i]))^2 + 2\sum_{i< j}w_iw_j((1-p)x[i] + pr(x[i]))((1-p)x[j] + pr(x[j])).
\end{align*}

We take expectation with respect to the input $x$,
\begin{align*}
\mathbb{E}(X_\text{test}^2) &= \sum_{i=1}^d w_i^2\mathbb{E}((1-p)^2x[i]^2 + 2(1-p)px[i]r(x[i])+ p^2r(x[i])^2) + 2\sum_{i< j}w_iw_j \mathbb{E}(p^2r(x[i])r(x[j]))
\\ &= \sum_{i=1}^d w_i^2(\frac{1}{2}p^2 -p +1) + \frac{p^2}{\pi}\sum_{i< j}w_iw_j.
\end{align*}

Using the fact that $\text{Var}(X_\text{test}) = \mathbb{E}(X_\text{test}^2) - (\mathbb{E}(X_\text{test}))^2$, we can obtain that
\begin{align}
\text{Var}(X_\text{test}) = \sum_{i=1}^d w_i^2((\frac{1}{2} - \frac{1}{2\pi})p^2 -p +1).
\label{eq:xtest_var}
\end{align}

With Eqn.~\ref{eq:xtrain_var} and Eqn.~\ref{eq:xtest_var}, we have
\begin{align}
\frac{\text{Var}(X_\text{test})}{\text{Var}(X_\text{train})}= \frac{(\frac{1}{2} - \frac{1}{2\pi})p^2 -p +1}{1-\frac{1}{2}p - \frac{1}{2\pi}p^2}.
\label{eq:shift}
\end{align}
\end{proof}

In Eqn.~\eqref{formula}, the range of the shift ratio lies on the interval $[0.8,1]$. In particular, when $p = 0.95$, $\text{Var}(X_\text{test})/\text{Var}(X_\text{train}) \approx 0.9377$, therefore $\text{Var}(X_\text{test})$ is close to $\text{Var}(X_\text{train})$. This shows that in Drop-Activation, the difference in the variance of inputs to a BN layer between the training and testing phase is rather minor.

We further demonstrate numerically that Drop-Activation does not generate an enormous shift in the variance of the internal covariates when going from the training time to the testing time. We train ResNet164 with CIFAR100. ResNet164 consists of a stack of three stages. Each stage contains 54 convolution layers with the same spatial size.  %\textbf{YK: I don't know what this means}.
We observe the statistics of the output of the second stage by evaluating its shift ratio. We compute the variances of the output for each channel and then average the channels' variance. As shown in Figure~\ref{fig:var_shift}, the shift ratio stabilizes close to $1$ at the end of the training, which is consistent with our analysis.

In summary, by maintaining the statistical property of the internal output of hidden layers in testing time, Drop-Activation can be combined with BN to improve performance.

\section{Experiments}
\label{sec:4}
In this section, we empirically evaluate the performance of Drop-Activation and demonstrate its effectiveness. We apply Drop-Activation to modern deep neural architectures on various datasets. This section is organized as followed. Section \ref{subsec:design} contains basic experiment settings. In Section \ref{subsec:dataset}, we introduce the datasets and implementation details. In section \ref{subsec:result}, we present the numerical results.

\subsection{Experiment design}
\label{sec:exp design}
Our experiments are to demonstrate the following points:
\textbf{(1) Comparison with RReLU:} Due to the similarity between the activation function used in our proposed method when having $f$ as ReLU in Eqn. \eqref{eq:output-L-da-test} and the randomized leaky rectified linear units (RReLU), one may speculate that the use of RReLU gives similar performance. We show that this is indeed not the case by comparing Drop-Activation with the use of RReLU.
\textbf{(2) Improvement upon modern neural network architectures:} We show the improvement that Drop-Activation brings is rather universal by applying it to different modern network architectures on a variety of datasets.
\textbf{(3) Compatibility with other approaches:} We show that Drop-Activation is compatible with other popular regularization methods by combining them in different network architectures.

%We design our experiments as follows. We are going to show that our method is better than the existing method that is similar to ours. As a consequence, we compare Drop-Activation with randomized leaky rectified linear units (RReLU). Then we apply Drop-Activation to different modern models and different datasets to show its effectiveness to improve generalization and work well with modern architectures. In the end, we combine Drop-Activation with some other popular regularization methods to show Drop-Activation can be compatible with them.

\label{subsec:design}
\textbf{Comparison with RReLU:} RReLU is proposed in \cite{rrelu} with the following training scheme for an input vector $x$,
\begin{equation}
\text{RReLU}(x)[j] = \left\{
\begin{aligned}
x[j] ,&\quad\ x[j]\geq 0,\\
U_jx[j],&\quad\ x[j]<0,
\end{aligned}
\right.
\label{eq:rrelu}
\end{equation}
where $U_j$ is a random variable with a uniform distribution $\mathcal{U}(a, b)$ with $0<a<b<1$. In the case of ReLU in Drop-Activation, a comparison between Eqn.~\eqref{eq:droprelu} with Eqn.~\eqref{eq:rrelu} shows that the main difference between our approach and RReLU is the random variable used on the negative axis. It can be seen from Eqn. \eqref{eq:rrelu} that RReLU passes the negative data with a random shrinking rate, while Drop-Activation randomly lets the complete information pass. The parameters $a$ and $b$ in RReLU are set at 1/8 and 1/3 respectively, as suggested in \cite{rrelu}.

\textbf{Improvement upon modern neural network architectures:} The residual-type neural network structures greatly facilitate the optimization for deep neural network \cite{resnet} and are employed by ResNet \cite{resnet}, PreResNet \cite{preresnet}, DenseNet \cite{densenet}, ResNeXt \cite{resnext}, WideResNet (WRN)\cite{wrn} and SENet \cite{se}. We demonstrate that Drop-Activation works well with these modern architectures. Moreover, since these networks use BN to accelerate training and may contain Dropout to improve generalization. e.g., WRN, these experiments also show the ability of Drop-Activation to work in synergy with the prevalent training techniques.

\textbf{Compatibility with other regularization approaches:} To further show that Drop-Activation can cooperate well with other training techniques, we combine Drop-Activation with two other popular data augmentation approaches: Cutout \cite{cutout} and AutoAugment \cite{autoaugment}. Cutout randomly masks a square region of training data and AutoAugment uses reinforcement learning to obtain an improved data augmentation scheme.

\subsection{Datasets and implementation details}
\label{subsec:dataset}
\begin{figure}
\centering
%\subfloat[]{
\begin{minipage}[t]{0.4\textwidth}
\vspace{0pt}
\includegraphics[width=1\textwidth]{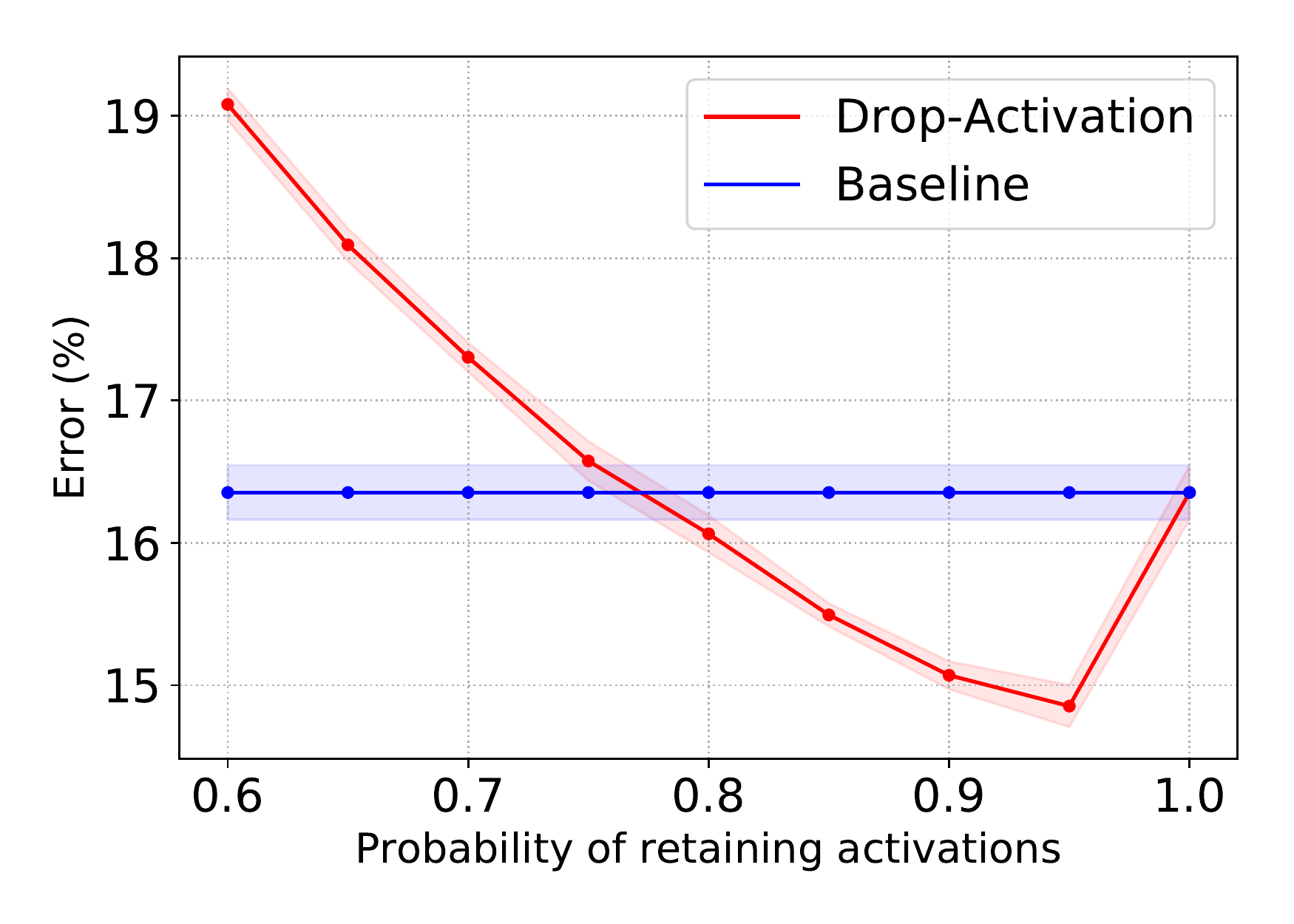}
\caption{Validation error on CIFAR10 with 95\% confidence intervals with respect to the probability $p$ of retaining activation (average of $20$ runs).}
\label{fig:parameters}
\end{minipage}
\hspace{.3in}
%\subfloat[]{
\begin{minipage}[t]{0.4\textwidth}
\vspace{0pt}
\includegraphics[width=1\textwidth]{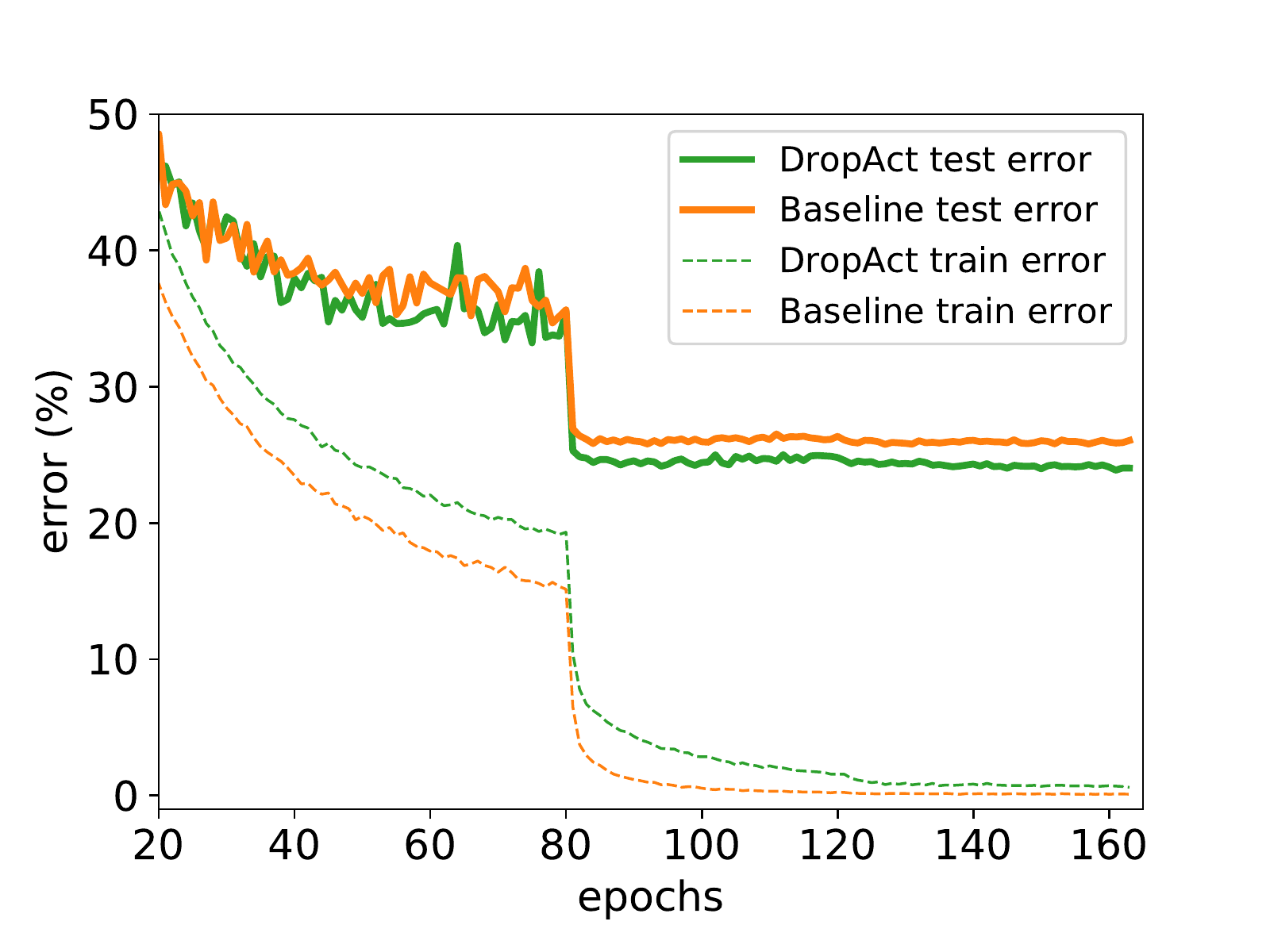}
\caption{Training curves on CIFAR100 with ResNet164.}
\label{fig:resnet-164}
\end{minipage}

\vspace{-0.4cm}
\end{figure}

\textbf{Choosing the probability of retaining activation:}
In our method, the only parameter that needs to be tuned is the probability $p$ of retaining activation. To get a rough estimate of what $p$ is, we train a simple network on CIFAR10 without data augmentation and perform a grid search for $p$ on the interval $[0.6, 1.0]$, with a step size equal to $0.05$. The simple network consists of three convolution layers and two fully connected layers, and details are in the Appendix. We split the train set of CIFAR10 into two parts, 10\% for validation and 90\% for training. The Figure~\ref{fig:parameters} shows the validation error on CIFAR10 versus $p$, which is minimal at $p=0.95$. Each data point is averaged over the outcomes of $20$ trained neural-networks. Based on this observation, we choose $p=0.95$ for all experiments.

\noindent\textbf{Datasets and implementation:} We train the models with Drop-Activation on CIFAR10, CIFAR100 \cite{cifar}, SVHN \cite{svhn}, EMNIST (``Balanced'') \cite{emnist} and ImageNet 2012 \cite{ImageNet} (random cropping size 224$\times$224). When applying Drop-Activation to these models, we directly substitute all the original ReLU function with Drop-Activation except for the case of ImageNet. In particular, due to the relatively underfitting of training on ImageNet, only ReLUs in the last two stages of networks are modified by Drop-Activation. All the models are optimized using SGD with a momentum of $0.9$ \cite{Sutskever:2013}. The other implementation details are given in the Appendix.

\subsection{Experiment results}
\label{subsec:result}
Table~\ref{tab:c100}, \ref{tab:svhnandemnist} and \ref{tab:imagenet} show the testing error on different datasets. The baseline results are from original networks without Drop-Activation. Table~\ref{tab:trainingtime} shows the training time of different models. In what follows, we discuss how our results support the points raised in Section \ref{sec:exp design} and analyse the training time of applying Drop-Activation.

\textbf{Comparison with RReLU: } As shown in Table~\ref{tab:c100}, RReLU may have worse performance than the baseline method. However, Drop-Activation consistently results in superior performance over RReLU and almost all baseline networks. Although Drop-Activation can not reduce the testing error of ResNeXt-8$\times$64d on CIFAR10, Drop-Activation with DenseNet190-40 has the best testing error smaller than that of the original ResNeXt29-8$\times$64d.

\begin{table}[htbp]
\small
  \centering
  
    \begin{tabular}{|c|c|c|c|c|c|c|}
    \toprule
          & \multicolumn{3}{c|}{CIFAR10} & \multicolumn{3}{c|}{CIFAR100} \\
\cmidrule{2-7}          & Baseline & RReLU & Drop-Act & Baseline & RReLU & Drop-Act \\
    \midrule
    VGG19(BN) & 6.56$\pm$0.13  & 6.60$\pm$0.22  & \textbf{6.38$\pm$0.07}  & 28.67$\pm$0.30     & 28.62$\pm$0.15     & \textbf{28.55$\pm$0.28} \\
    ResNet110 & 6.77$\pm$0.27  & 7.37$\pm$0.22  & \textbf{6.25$\pm$0.06}  & 28.24$\pm$0.13     & 29.64$\pm$0.06     & \textbf{27.91$\pm$0.18} \\
    ResNet164 & 5.94$\pm$0.27     & 6.08$\pm$0.03     & \textbf{5.62$\pm$0.08}     & 25.86$\pm$0.42  & 24.78$\pm$0.43  & \textbf{24.18$\pm$0.22}  \\
    PreResNet164 & 5.01$\pm$0.03  & 5.17$\pm$0.12  & \textbf{4.87$\pm$0.16}  & 23.49$\pm$0.17  & 23.21$\pm$0.05  & \textbf{22.79$\pm$0.16}  \\
    WideResNet28-10 & 3.85$\pm$0.13  & 4.32$\pm$0.05  & \textbf{3.74$\pm$0.05}  & 18.84$\pm$0.26  & 19.53$\pm$0.13  & \textbf{18.14$\pm$0.22}  \\
    DenseNet100-12 & 4.73$\pm$0.10  & 5.06$\pm$0.03  & \textbf{4.38$\pm$0.09}  & 22.66$\pm$0.25  & 22.59$\pm$0.20  & \textbf{21.80$\pm$0.21}  \\
    DenseNet190-40 & 3.91$\pm$0.15  &   3.84$\pm$0.08    & \textbf{3.51$\pm$0.06}  & 17.28$\pm$0.45   &   18.58$\pm$0.12    & \textbf{16.80$\pm$0.12}  \\
    ResNeXt29-8$\times$64 & 3.95$\pm$0.05  & 4.56$\pm$0.16  & 3.95$\pm$0.18  & 18.56$\pm$0.38  & 18.65$\pm$0.08  & \textbf{17.65$\pm$0.16}  \\
    \bottomrule
    \end{tabular}%

  \caption{Test error (\%) on CIFAR10 an CIFAR100. The test accuracy is averaged over three repeated experiments. We use Baseline to indicate the usage of the original architecture without modifications.}
  \label{tab:c100}
  \vspace{-0.5cm}
\end{table}%

\textbf{Application to modern models:} As shown in Table~\ref{tab:c100}, Drop-Activation in almost all cases improves the testing accuracy consistently comparing to Baseline for CIFAR10 and CIFAR100. To further demonstrate this, we apply Drop-Activation to various neural-network architectures and demonstrate the successes on the datasets SVHN, EMNIST, and ImageNet. Again, in Table~\ref{tab:svhn_minist} and \ref{tab:imagenet} we see a consistent improvement when Drop-Activation is used. %{\color{blue}\sout{In particular, Drop-Activation improves ResNet, PreResNet, and WRN by reducing the relative test error for CIFAR10, CIFAR100 or SVHN by over 3.5\%.}}

Therefore, Drop-Activation can work with most modern networks for different datasets. Besides, our results implicitly show that Drop-Activation is compatible with regularization techniques such as BN or Dropout used in training these networks.

\begin{table}[htbp]
\begin{minipage}[t]{0.55\linewidth}
\small
  \centering
    \begin{tabular}{|c|c|c|c|c|}
    \toprule
    Models & \multicolumn{2}{c|}{SVHN} & \multicolumn{2}{c|}{EMNIST} \\
\cmidrule{2-5}          & Base & Drop-Act & Base & Drop-Act \\
    \midrule
    ResNet164 & -     & -     & 8.85  & \textbf{8.82}  \\
    PreResNet164 & -     & -     & 8.88  & \textbf{8.72}  \\
    WRN16-8 & 1.54  & \textbf{1.46}  & -     & - \\
    WRN28-10 & -     & -     & 8.97  & \textbf{8.72}  \\
    DenseNet100-12 & 1.76  & \textbf{1.71}  & \textbf{8.81}  & 8.90  \\
    ResNeXt29,8*64 & 1.79  & \textbf{1.69}  & 9.07  & \textbf{8.91}  \\
    \bottomrule
    \end{tabular}%

  \caption{Test error (\%) on SVHN, EMNIST (Balanced). The Baseline results of WRN and DenseNet for SVHN are obtained from the original papers.}\label{tab:svhn_minist}
  \label{tab:svhnandemnist}%
\end{minipage}
\begin{minipage}[t]{0.45\linewidth}

% Table generated by Excel2LaTeX from sheet 'Sheet3'
\small
  \centering

    \begin{tabular}{|c|c|c|}
    \toprule
          & \multicolumn{2}{c|}{ImageNet 2012} \\
\cmidrule{2-3}      Models    & Baseline & Drop-Act \\
    \midrule
    ResNet34 & 26.07  & \textbf{25.85}  \\
    SENet50 & 23.39  & \textbf{23.18}  \\
    \bottomrule
    \end{tabular}%

  \caption{Validation error (\%) on ImageNet.}
 \label{tab:imagenet}%

\end{minipage}
\end{table}%

% \begin{table}[htbp]
% \small
%   \centering
%     \begin{tabular}{|c|c|c|c|c|c|c|}
%     \toprule
%           & \multicolumn{2}{c|}{SVHN} & \multicolumn{2}{c|}{EMNIST} & \multicolumn{2}{c|}{ImageNet 2012} \\
% \cmidrule{2-7}          & Baseline & Drop-Act & Baseline & Drop-Act & Baseline & Drop-Act \\
%     \midrule
%     ResNet-164 & -     & -     & 8.85  & 8.82  & -     & - \\
%     PreResNet-164 & -     & -     & 8.88  & 8.72  & -     & - \\
%     WideResNet16-8 & 1.54  & 1.46  & -     & -     & -     & - \\
%     WideResNet28-10 & -     & -     & 8.97  & 8.72  & -     & - \\
%     DenseNet100-12 & 1.76  & 1.71  & 8.81  & 8.90  & -     & - \\
%     ResNeXt29-8*64 & 1.79  & 1.69  & 9.07  & 8.91  & -     & - \\
%     ResNet34 & -     & -     & -     & -     & 26.07  & 25.85  \\
%     SeNet50 & -     & -     & -     & -     & 23.39  & 23.18  \\
%     \bottomrule
%     \end{tabular}%
%     \caption{Test error (\%) on SVHN, EMNIST (Balanced) and ImageNet 2012. The Baseline results of WideResNet and DenseNet for SVHN are quoted from the original papers. The Baseline results for EMNIST and ImageNet and the Baseline result of ResNeXt for SVHN were generated by ourselves.}
%   \label{tab:svhn_emnist}%
%   \vspace{-0.5cm}
% \end{table}%

\textbf{Compatibility with other regularization approaches:} We apply Drop-Activation to network models that use Cutout or AutoAugment. As shown in Table~\ref{tab:combination}, Drop-Activation can further improve with Cutout or AutoAugment by decreasing the test error on CIFAR100 and CIFAR10.
\begin{table}[htbp]
\small
  \centering
  
   \begin{tabular}{|c|c|c|c|c|c|c|c|}
    \toprule
    \multirow{2}[4]{*}{Model} & \multirow{2}[4]{*}{Dataset} & \multirow{2}[4]{*}{Baseline} & \multirow{2}[4]{*}{DA} & \multicolumn{2}{c|}{with Cutout (CO)} & \multicolumn{2}{c|}{with AutoAug (AA)} \\
\cmidrule{5-8}          &       &       &       & CO    & CO+DA & AA    & AA+DA \\
    \midrule
    ResNet18 & CIFAR100 & 22.46  &   21.61    & 21.96  & 20.99  & -     & - \\
    ResNet164 & CIFAR100 & 25.86  & 24.18  & -     & -     & 21.12  & 20.39  \\
    WideResNet28-10 & CIFAR100 & 18.84  & 18.14  & 18.41  & 17.86  & 17.09  & 16.20  \\
    \midrule
    DenseNet190-40 & CIFAR10 & 3.91  & 3.51  & 3.15  & 2.79  & 2.54  & 2.36  \\
    \bottomrule
    \end{tabular}%

    \caption{Test error~(\%) for CIFAR100 or CIFAR10 with combination of Drop-Activation (DA) and Cutout (CO) or AutoAugement (AA). The results of Cutout are quoted from \cite{cutout}. The WideResNet result of AutoAug is quoted from \cite{autoaugment}.}
  \label{tab:combination}%
  \vspace{-0.5cm}
\end{table}%

\textbf{Training time:}
The increment of the computational cost of the Drop-Activation network compared with the ReLU network comes from the different realizations of Bernoulli random variables for each activation function. This results in an unavoidable increment of training time. Table~\ref{tab:trainingtime} shows the training time of each batch for different models. In particular, RReLU that we use is Pytorch official function. We train ResNet164 and WideResNet28 with batch size 128 on CIFAR10 using the workstation with CPU AMD Ryzen Threadripper 1920X and 2 GPUs 2080Ti. From Table~\ref{tab:trainingtime}, we can see that both Drop-Activation and officially implemented RReLU suffer from the training time increment.
\begin{table}[htbp]
\small
  \centering
  
    \begin{tabular}{|c|c|c|c|}
    \toprule
    Model & Baseline & RReLU & Drop-Activation \\
    \midrule
    ResNet164 & 0.151 & 0.175 & 0.223 \\
    WideResNet28-10 & 0.128 & 0.212 & 0.179 \\
    \bottomrule
    \end{tabular}%
   \caption{The training time (sec) for each batch of different models.}
   \label{tab:trainingtime}%
\end{table}%

\section{Conclusion}
In this paper, we propose Drop-Activation, a regularization method that introduces randomness on the activation function. Drop-Activation works by randomly dropping the nonlinear activations in the network during training and uses a deterministic network with modified nonlinearities for prediction.

The advantage of the proposed method is two-fold. Firstly, Drop-Activation provides a simple yet effective method for regularization, as demonstrated by the numerical experiments. Furthermore, this is supported by our analysis in the case of one hidden-layer. We show that Drop-Activation gives rise to a regularizer that penalizes the difference between nonlinear and linear networks. Future direction includes the analysis of Drop-Activation with more than one hidden layer. Secondly, experiments verify that Drop-Activation improves the generalization in most modern neural networks and cooperates well with some other popular training techniques. Moreover, we show theoretically and numerically that Drop-Activation maintains the variance during both training and testing time, and thus Drop-Activation can work well with Batch Normalization. These two properties should allow the wide applications of Drop-Activation in many network architectures.

\section{Conflict of Interest}
On behalf of all authors, the corresponding author states that there is no conflict of interest.

\vspace{1cm}
{\bf Acknowledgments.} S. Liang and H. Yang gratefully acknowledge the support of National Supercomputing Center (NSCC) Singapore \cite{nscc} and High-Performance Computing (HPC) of the National University of Singapore for providing computational resources, and the support of NVIDIA Corporation with the donation of the Titan Xp GPU used for this research. H. Yang was partially supported by National Science Foundation under the grant award 1945029.

\section{Appendix}
\subsection{The simple model for finding the best parameter $p$}
\begin{figure}[ht]
  \centering
  % Requires \usepackage{graphicx}
  \includegraphics[width=0.4\textwidth]{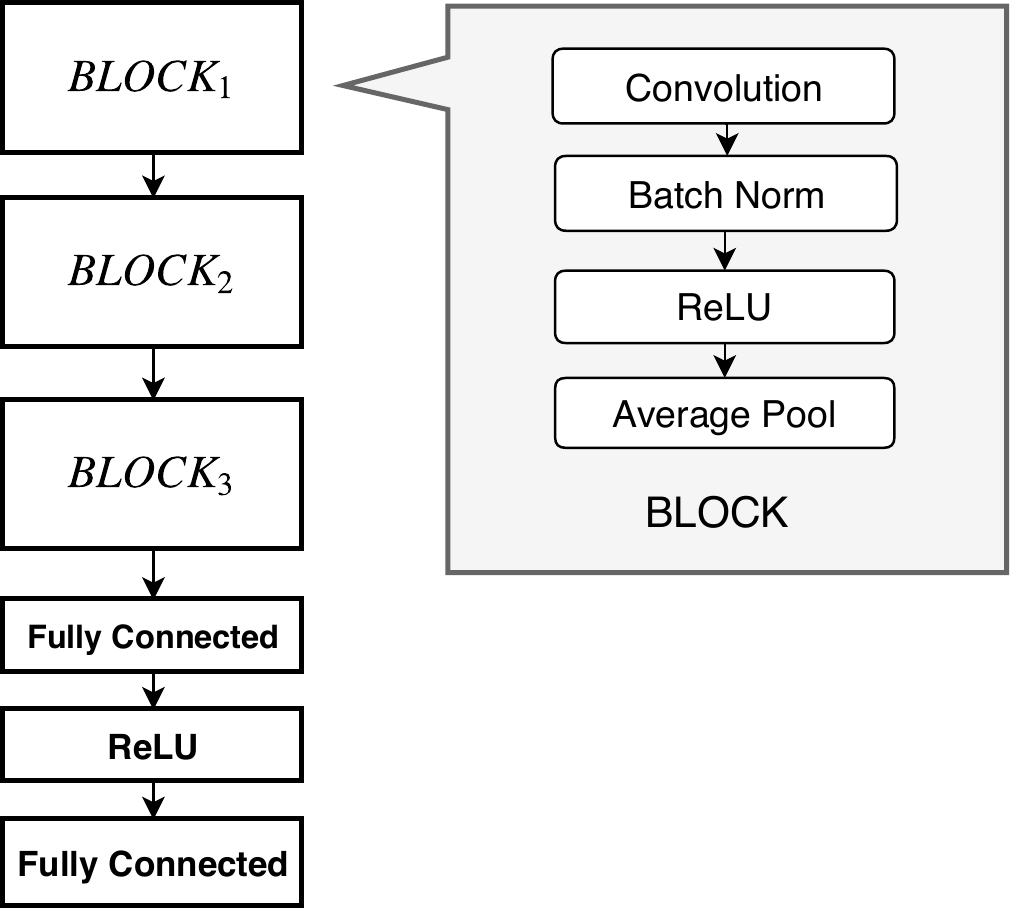}\\
  \caption{The model for finding the best parameter for Drop-Activation.}\label{fig:simplemodel}
\end{figure}

To find the best parameter for Drop-Activation, we perform a grid search on a simple model. The simple network consists of the following layers: We first stack three blocks, and each block contains convolution with $3\times3$ filter, BN, ReLU, and average pooling, as shown in Figure~\ref{fig:simplemodel}. The number of $3\times 3$ filters for $\text{Block}_1$, $\text{Block}_2$, $\text{Block}_3$ is 32, 64, 128 respectively. The widths for fully connected layers are 1000 and 10 respectively.

\subsection{Introdcution of datasets}
We use the following datasets in our numerical experiments,

\textbf{CIFAR:} Both CIFAR10 and CIFAR100 contain 60k color nature images of size 32 by 32. There are 50k images for training and 10k images for testing. CIFAR-10 has ten classes of objects and 6k for each class. CIFAR100 is similar to CIFAR10, except that it includes 100 classes and 600 images for each class. Normalization and standard data augmentation (random cropping and horizontal flipping) are applied to the training data as \cite{resnet}.
% For CIFAR-10, we train the models ResNet-110, PreResNet-164, DenseNet-BC-100-12, DenseNet-BC-190-40, ResNeXt-8$\times$64d, WideResNet-28-10. For CIFAR-100, the models that we train are the same as in the case for CIFAR-10 except that ResNet-110 is replaced with ResNet-164 using the bottleneck layer in \cite{preresnet}. We use the same hyper-parameters as in the original papers except that the batch-size for DenseNet-BC-190-40 is set to $32$. The models are optimized using SGD with a momentum of $0.9$ \cite{Sutskever:2013}.

\textbf{SVHN:} The dataset of Street View House Numbers (SVHN) contains ten classes of color digit images of size 32 by 32. There are about 73k training images, 26k testing images, and additional 531k images. The training and additional images are used together for training, so there are totally over 600k images for training. An image in SVHN may contain more than one digit, and the recognition task is to identify the digit in the center of the image. We preprocess the images following \cite{wrn}. The pixel values of the images are rescaled to $[0,1]$, and no data augmentation is applied.

\textbf{EMNIST:} EMNIST is a set of $28\times 28$ grayscale images containing handwritten English characters and digits. There are six different splits in this dataset and we use the split ``Balanced''. In the ``Balanced'' split, there are 131,600 images in total, including 112,800 for training and 18,800 for testing.
%Some pairs of characters with a similar shape regarding upper and lower cases belong to the same class, such as 'C' and 'c.'\textbf{YK: Is this detail necessary?}
% There are 47 distinct classes. For this classification task, we train the models ResNet-164, PreResNet-164, WideResNet-20-10, DenseNet-BC-100-12, and ResNeXt-8$\times$64d using the hyper-parameter settings for training CIFAR-100 in \cite{resnet, preresnet, wrn, densenet, resnext} respectively.

\textbf{ImageNet 2012:} The ImageNet 2012 dataset consists of 1.28 million training images and 50K validation images from 1,000 classes. The models are evaluated on the validation set. %{\color{blue}\sout{Due to the relatively underfitting of training on ImageNet, we only apply Drop-Activation to the last two stages of networks.}}
We train the models for 120 epochs with an initial learning rate 0.1.
\subsection{Implementation detail}
The hyper-parameters for different networks are shown in Table~\ref{tab:cifar-params}, \ref{tab:imgnet} and \ref{tab:svhn}, and we offer the explanation of hyper-parameter names in Table~\ref{tab:explain}.
\begin{table}[htbp]
\scriptsize
  \centering
    \begin{tabular}{|c|c|c|c|c|c|c|c|}
    \toprule
          & ResNet & PreResNet & WRN-28 & ResNext29-8*64 & VGG19(BN) & DenseNet190 & DenseNet100 \\
    \midrule
    Batch size & 128   & 128   & 128   & 128   & 128   & 32    & 64 \\
    Epoch & 164   & 164   & 200   & 300   & 200   & 300   & 300 \\
    Optimizer & SGD(0.9) & SGD(0.9) & SGD(0.9) & SGD(0.9) & SGD(0.9) & SGD(0.9) & SGD(0.9) \\
    Depth & -     & -     & 28    & 29    & 19    & 190   & 100 \\
    Schedule & 81/122 & 81/122 & 80/120/160 & 150/225 & 80/140 & 150/225 & 150/225 \\
    Weight-decay & 1.00E-04 & 1.00E-04 & 5.00E-04 & 5.00E-04 & 1.00E-04 & 1.00E-04 & 1.00E-04 \\
    Gamma & 0.1   & 0.1   & 0.2   & 0.1   & 0.1   & 0.1   & 0.1 \\
    Grow-rate & -     & -     & -     & -     & -     & 40    & 12 \\
    Widen-factor & -     & -     & 10    & 4     & -     & -     & - \\
    Cardinality & -     & -     & -     & 8     & -     & -     & - \\
    LR    & 0.1   & 0.1   & 0.1   & 0.1   & 0.1   & 0.1   & 0.1 \\
    Dropout & -     & -     & 0.3   & -     & -     & -     & - \\
    \bottomrule
    \end{tabular}%
  \caption{Hyper-parameter setting for training models on CIFAR10/100 and EMNIST.}
  \label{tab:cifar-params}%
\end{table}%

\begin{table}[htbp]
\scriptsize
\begin{minipage}[b]{0.4\linewidth}

  \centering
     \begin{tabular}{|c|c|c|}
    \toprule
          & ResNet34 & SENet50 \\
    \midrule
    Batch size & 256   & 256 \\
    Epoch & 120   & 120 \\
    Optimizer & SGD(0.9) & SGD(0.9) \\
    Depth & 34    & 50 \\
    Schedule & 30/60/90 & 30/60/90 \\
    Weight-decay    & 1.00E-04 & 1.00E-04 \\
    Gamma & 0.1   & 0.1 \\
    LR    & 0.1   & 0.1 \\
    \bottomrule
    \end{tabular}%
  \caption{Hyper-parameter setting for training models on ImageNet.}
  \label{tab:imgnet}%
\end{minipage}
\hspace{.5in}
\begin{minipage}[b]{0.5\linewidth}
% Table generated by Excel2LaTeX from sheet 'Sheet3'

  \centering
    \begin{tabular}{|c|c|c|c|}
    \toprule
          & WRN-16 & ResNext29-8*64 & DenseNet100 \\
    \midrule
    Batch size & 128   & 128   & 64 \\
    Epoch & 160   & 100   & 40 \\
    Optimizer & SGD(0.9) & SGD(0.9) & SGD(0.9) \\
    Depth & 16    & 29    & 100 \\
    Schedule & 80/120 & 40/70 & 20/30 \\
    Weight-decay & 5.00E-04 & 5.00E-04 & 1.00E-04 \\
    Gamma & 0.2   & 0.1   & 0.1 \\
    Grow-rate & -     & -     & 12 \\
    Widen-factor & 8     & 4     & - \\
    Cardinality & -     & 8     & - \\
    LR    & 0.01  & 0.1   & 0.1 \\
    Dropout & 0.4   & -     & - \\
    \bottomrule
    \end{tabular}%
  \caption{Hyper-parameter setting for training models on SVHN.}
 \label{tab:svhn}%
\end{minipage}
\end{table}%

\begin{table}[htbp]
		\small
		\centering
		\begin{tabular}{|l|l|}
			\toprule
			Batch size & Number of samples for training at each iteration \\
			Epoch & Number of total epochs to train \\
			Depth & The depth of network \\
			Schedule & Decrease learning rate at these epochs \\
			Weight-decay    & The coefficient of l2 loss \\
			Gamma & Learning rate is multiplied by Gamma on schedule \\
			Widen-factor & Widen factor \\
			Cardinality & Model cardinality (group) \\
			LR    & initial learning rate \\
			Dropout  & Dropout ratio \\
			\bottomrule
		\end{tabular}%
		\caption{The explanation of hyper-parameter names.}
        \label{tab:explain}%
		\vspace{-0.2cm}
	\end{table}%
\end{document}